\documentclass{article}
\usepackage{amsmath, amssymb, amsfonts}
\usepackage{algorithm}
\usepackage{algorithmic}
\usepackage{subfig}
\usepackage{graphicx}
\usepackage{booktabs}
\usepackage{authblk}
\usepackage{hyperref}
\usepackage{cite}
\usepackage{xcolor}
\usepackage{url}
\usepackage{adjustbox}
\usepackage{pdftexcmds}

\DeclareGraphicsExtensions{.pdf,.png,.jpg}
\begin{document}

\title{Adapting the Biological SSVEP Response to Artificial Neural Networks}

\author[1,2,*]{Emirhan Böge}
\author[2,3,*]{Yasemin Gunindi}
\author[2]{Erchan Aptoula}
\author[3]{Nihan Alp}
\author[2]{Huseyin Ozkan}

\affil[1]{School of Informatics, University of Edinburgh, Edinburgh, UK}
\affil[2]{Faculty of Engineering and Natural Sciences, VPAlab, Sabanci University, Istanbul, Turkiye}
\affil[3]{Faculty of Arts and Social Sciences, AlViNlab, Sabanci University, Istanbul, Turkiye}

\date{}

\maketitle

\renewcommand{\thefootnote}{}
\footnotetext{* These authors contributed equally to this work.}
\renewcommand{\thefootnote}{\arabic{footnote}}

\begin{abstract}
Neuron importance assessment is crucial for understanding the inner workings of artificial neural networks (ANNs) and improving their interpretability and efficiency. This paper introduces a novel approach to neuron significance assessment inspired by frequency tagging, a technique from neuroscience. By applying sinusoidal contrast modulation to image inputs and analyzing resulting neuron activations, this method enables fine-grained analysis of a network's decision-making processes. Experiments conducted with a convolutional neural network for image classification reveal notable harmonics and intermodulations in neuron-specific responses under part-based frequency tagging. These findings suggest that ANNs exhibit behavior akin to biological brains in tuning to flickering frequencies, thereby opening avenues for neuron/filter importance assessment through frequency tagging. The proposed method holds promise for applications in network pruning, and model interpretability, contributing to the advancement of explainable artificial intelligence and addressing the lack of transparency in neural networks. Future research directions include developing novel loss functions to encourage biologically plausible behavior in ANNs.
\end{abstract}

\section{Introduction}
\label{sec:intro}
Neuron importance assessment refers to quantifying the significance of individual neurons within an artificial neural network (ANN), with respect to their contribution to overall network performance and output. Such tools have a wide application range, mostly in the context of explainable artificial intelligence (XAI), that aims to improve the long-criticized lack of transparency of neural networks \cite{haar2023analysis}; and network pruning, where by removing less influential neurons, one can forge computationally less intensive models with often negligible performance loss \cite{he2022filter}.

The assessment of neuron significance within ANNs has been gaining popularity. Notable advances include layer-wise relevance propagation \cite{bach2015pixel}, which backpropagates output predictions throughout the network while assigning relevance scores to individual neurons. DeepLIFT \cite{shrikumar2017learning} on the other hand contrasts neuron activation with the final decision of the network based on a chosen reference input that is ``neutral''. This comparison enables the assignment of contribution scores to individual neurons. A neuron/filter pruning technique for convolutional neural networks (CNN) has been also reported \cite{molchanov2019importance} through the Taylor expansion of the loss function, approximating the elimination effect of an individual neuron on the model's loss. Another study \cite{he2019filter} has tackled the same issue through inter-filter mutual relationships. Moreover, the Neuron Shapley approach \cite{ghorbani2020neuron} relies on computing the average marginal neuron contributions across all possible neuron combinations, while Neuron-Level Plasticity Control \cite{paik2020overcoming} estimates a neuron's capacity to maintain previously learned knowledge during adaptation to new tasks with the end of significance quantification. More recently, the context of adversarial attacks has been explored \cite{zhang2022improving} through the activation and deactivation of individual neurons, so as to assess which among them influence a model's vulnerability.

In contrast to the aforementioned studies, this paper proposes a radically different approach to neuron significance assessment, based on the concept of frequency tagging, adopted from the neuroscientific investigation of the biological brain \cite{retter2021harmonic}. To the best of our knowledge, it has never been explored before in the context of artificial neural networks.
\begin{figure*}[ht]
  \begin{center}
  \includegraphics[width=0.8\textwidth]{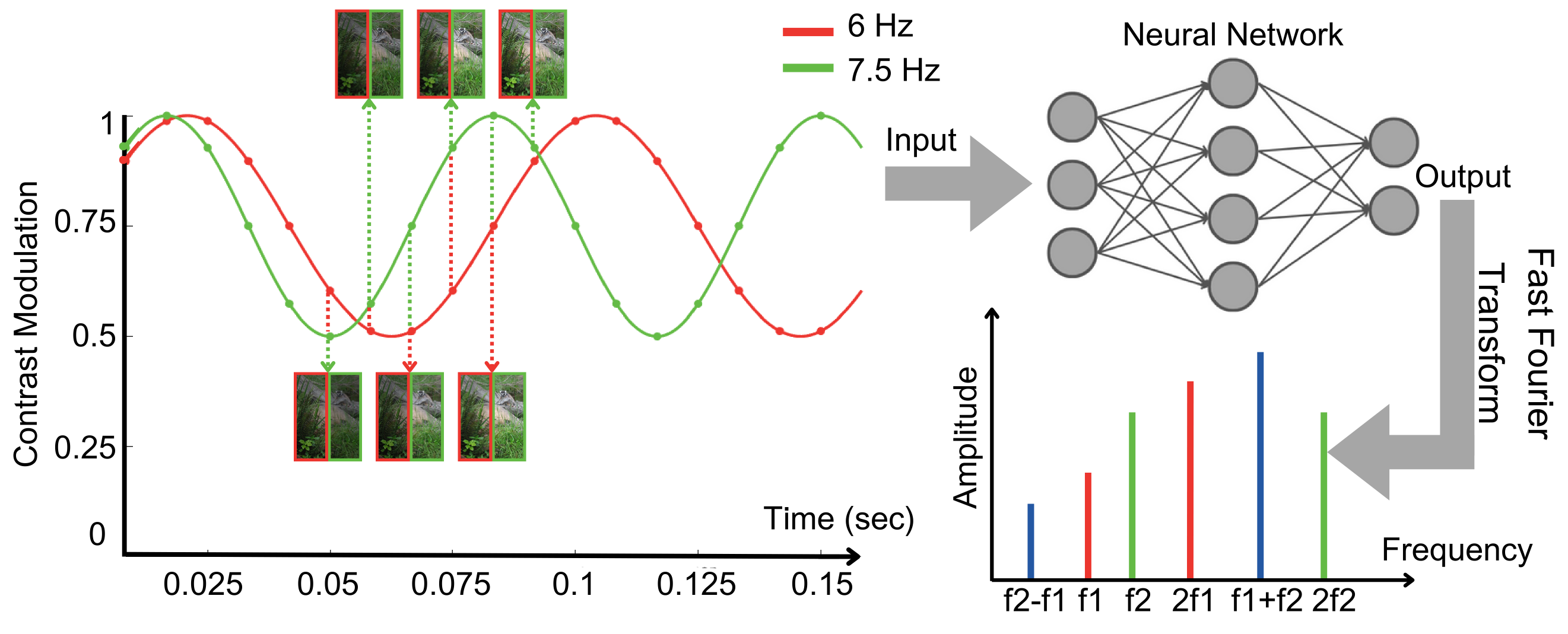}
  \end{center}
  \caption{The proposed frequency tagging approach for a color image. Its left half is tagged (i.e.~contrast-modulated) with 6 Hz whereas the right half is tagged with 7.5 Hz. The tagged images are sequentially fed into an arbitrary (convolutional) neural network, in the order of their tagging, and the neuron's activation responses are acquired across time. The temporal sequence is then transformed into the frequency domain, with symbolic amplitudes illustrating the network's response strength at specific frequencies.}
  \label{process}
\end{figure*}
More specifically, if the human brain is presented with a visual stimulus flickering (i.e.~changing intensity regularly) at a specific frequency, then the neural response gets tuned to the harmonics of the same frequency. This is a well-established observation in neuroscience literature \cite{boremanse2013objective, norcia2015steady,alp2022neural}, and can be directly observed from an electroencephalogram (EEG). Frequency tagging (Fig.~\ref{process}) is a method of artificially flickering parts of a visual stimulus with distinct frequencies (i.e.~so parts are ``tagged") with the aim of isolating the traces of part-based processing in the brain \cite{norcia2015steady}. 

In this paper, it has been hypothesized that one should observe the same effect in an ANN as well, if it indeed mimics a biological brain as often claimed \cite{wang2021triple}. To test this, we have focused on a prominent neuroscientific experimental paradigm \cite{boremanse2013objective}, and replaced the biological brain with an ANN. The specifics of frequency tagging and its adaptation to ANNs as well as the tools that can be

 used for node significance assessment are presented next in Section \ref{sec:approach}. Then the proof-of-concept experiments (Section \ref{section.implementation}) that have been conducted with color images and a Resnet-32 CNN, with the end of observing notable harmonics under part-based frequency tagging are described. The paper ends with a discussion of our results and various deep learning applications that can benefit from the proposed neuron significance assessment method (Section \ref{sec:conclusion}).

\section{Frequency tagging for artificial neural networks}
\label{sec:approach}
The frequency tagging technique involves modulating the contrast of a stimulus at specific frequencies over time \cite{regan1966effect}. In more detail, various parts of an image, commonly the left and right sides \cite{boremanse2013objective}, undergo sinusoidal contrast modulation and therefore ``flicker'' at specific frequencies. The objective of introducing these contrast-modulated input stimuli to biological brains is to evoke steady (non-transient) neural responses, known as \textit{steady-state visually evoked potentials} (SSVEPs) \cite{regan1969clinical}. 

First reported by Adrian and Matthews \cite{adrian1934berger}, as ``a series of potential waves having the same frequency as that of the flicker'', steady-state responses to visual stimuli allow discernible brain signals to be tracked. These signals stand apart from other concurrent endogenous processes, allowing researchers to isolate underlying neural correlates of stimulus properties. Consequently, by contrast-modulating different parts of visual stimuli, the contribution of these parts to the signals can be deciphered through their given frequencies. 

Inspired by face perception studies in neuroscience and computer vision \cite{boremanse2013objective,kanwisher2006fusiform,rossion2000n170,alp2022neural}, this paper proposes to apply frequency tagging to digital color images with two different frequencies and then provide them as input to a CNN, to assess based on the aforementioned principle, the importance/utility of individual neurons/filters. 

In more detail, to emulate the frequency tagging process, the pixel values of each channel of a digital color image are independently multiplied by a scalar coefficient $\sin(\omega_i)$ derived from a sinusoid of frequency $f$. This coefficient is scaled into the range $[0.5,1]$ to prevent excessive loss of luminance (see Fig.~\ref{process}). The angles $\omega_i$ are computed as: 
\begin{equation}
     \omega_i = 2\pi \times f \times i / \text{FPS} + \phi.
\end{equation}
The phase $\phi$ is set to $0$, and frames per second (FPS) is fixed at 120. With a duration of 2 seconds, this results in the creation of 240 color images, where $i \in \{0,\dots, 239\}$, collectively forming the tagged version of the original color input image. In our setting, this process is applied independently to the left and right halves of a given image using respectively $f_{left}=6$ Hz and $f_{right}=7.5$ Hz \cite{alp2022neural}. At this point it is critical that the resulting images are provided as input to the network in the same order that they have been produced. Consequently, one can at this stage collect the neuron activations corresponding to each of the contrast modulated input images, thus obtaining the SSVEPs in the form of a sequence of 240 numerical activations per neuron (or filter, in the case of CNNs). If the output of a neuron is non-scalar, it can be converted to be so, for instance by taking the average or maximum of a filter output in the case of CNNs.

\begin{figure*}[ht]
  \begin{center}
  \includegraphics[width=0.9\textwidth]{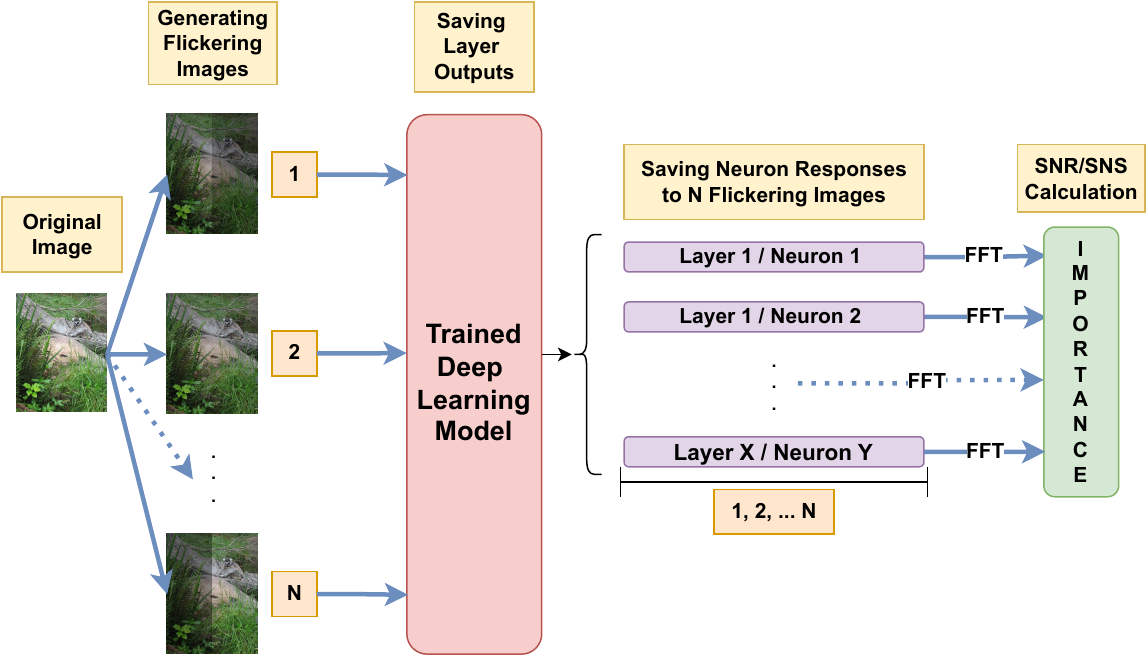}
  \end{center}
  \caption{Illustration of the proposed neuron significance assessment process.}
  \label{flowchart}
\end{figure*}

When the brain processes a flickering stimulus that is contrast-modulated as described, it operates as a nonlinear biological system, producing not only the given frequencies in the input (i.e.~fundamental frequencies: $f_1, f_2$) but also the harmonics (i.e.~integer multiples of the fundamental frequencies: $nf_1, mf_2$) \cite{regan1988frequency}. Moreover, due to the brain's non-isolated processing of different parts, intermodulation (IM) components do also emerge, defined as $nf_1 \pm mf_2$. Given the inherent nonlinearity of neural networks, like brain, we anticipate observing both harmonics and IMs in our setting as well.


\begin{figure}[t]
  \begin{center}
  \includegraphics[width=0.42\textwidth]{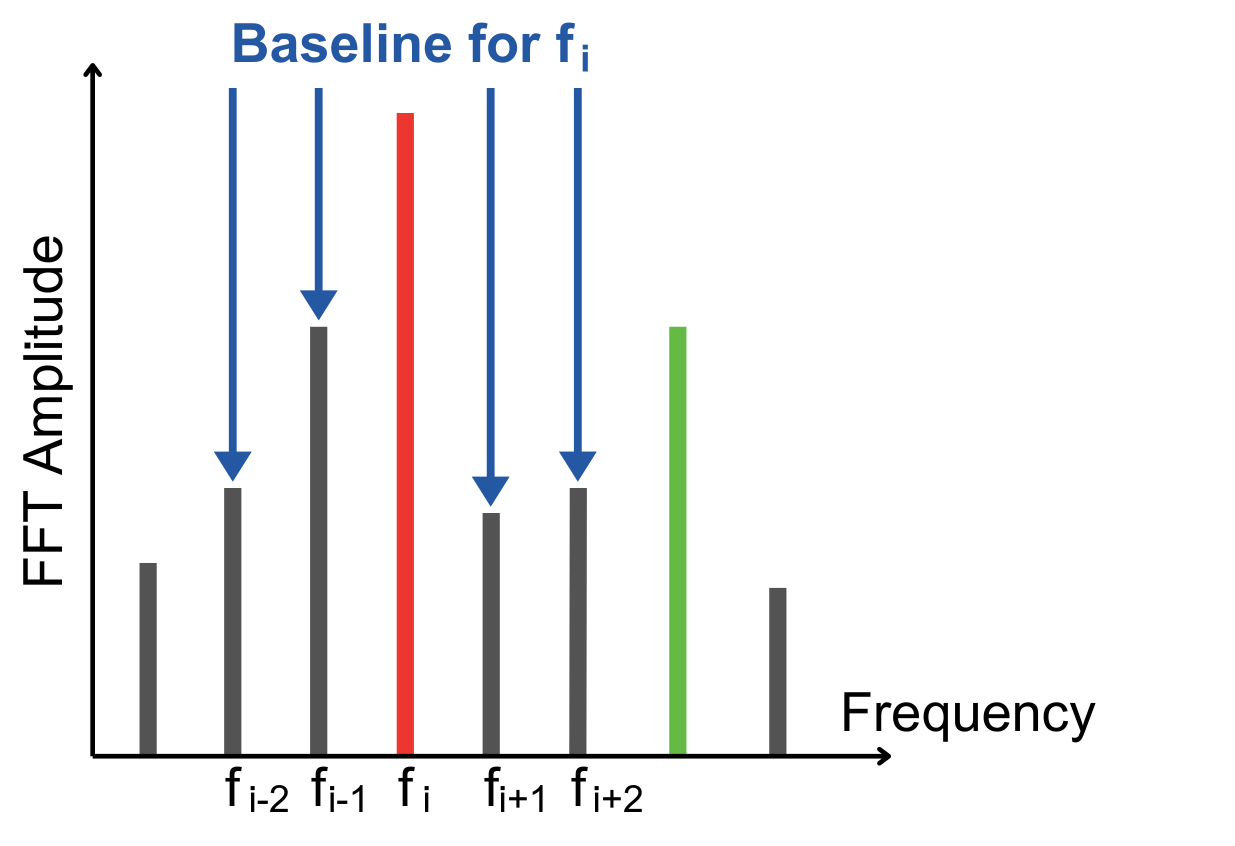}
  \end{center}
  \caption{Illustration of the frequencies employed for SNR calculation. Baseline = Mean of Amplitudes $\{f_{i-2},f_{i-1}, f_{i+1}, f_{i+2}\}$.
  SNR of $f_i$ = Amplitude of $f_i$ / Baseline.}
  \label{fig:snr}
\end{figure}

Now that we have access to the sequences of neuron activations across time (i.e.~SSVEPs), the next step consists in determining which of them exhibit the modulation frequencies of the inputs (in other words, which neurons behave more similarly to their biological counterparts and ergo are more important). To this end, we operate in the frequency domain via the Fast Fourier Transform (FFT), and rely on standard tools such as Signal-to-Noise Ratio (SNR) (Signal-to-Noise-Subtracted-Amplitudes (SNS) constitute an alternative to SNR encountered in the neuroscience state-of-the-art)\cite{norcia2015steady,retter2021harmonic}. More precisely, as far as SNR is concerned, each frequency $\nu$ in the FFT output of each SSVEP is controlled in terms of importance, by dividing its amplitude with the average amplitude of its surrounding frequencies (i.e.~the baseline, commonly a couple of neighboring frequencies from the left and right of $\nu$). Thus, an amplitude greater than its neighbors, i.e., a frequency with high SNR, is going to stand out. It is essential to carefully define the range of the baseline to ensure that fundamental frequencies are excluded, preventing them from being treated as noise (Fig.~\ref{fig:snr}). For a discrete detection of important nodes, a threshold across the calculated SNR values can be used.

Consequently, following the well-established neuroscientific principles regarding biological brains \cite{boremanse2013objective}, network neurons exhibiting high SNR and SNS values are considered crucial for the model’s functionality as they respond distinctly to the specific modulated frequencies of the network's input. 

\section{Implementation and findings}
\label{section.implementation}
As far as implementation is concerned, a ResNet-32 model \cite{he2016deep} trained with 50k samples of the CIFAR-10 dataset \cite{krizhevsky2009learning} has been employed. 100 randomly selected test images from the same dataset (non-overlapping with the training set) have been contrast modulated following the procedure outlined in Section \ref{sec:approach}. 
\begin{figure*}[ht]
  \begin{center}
  \includegraphics[width=1.0\textwidth]{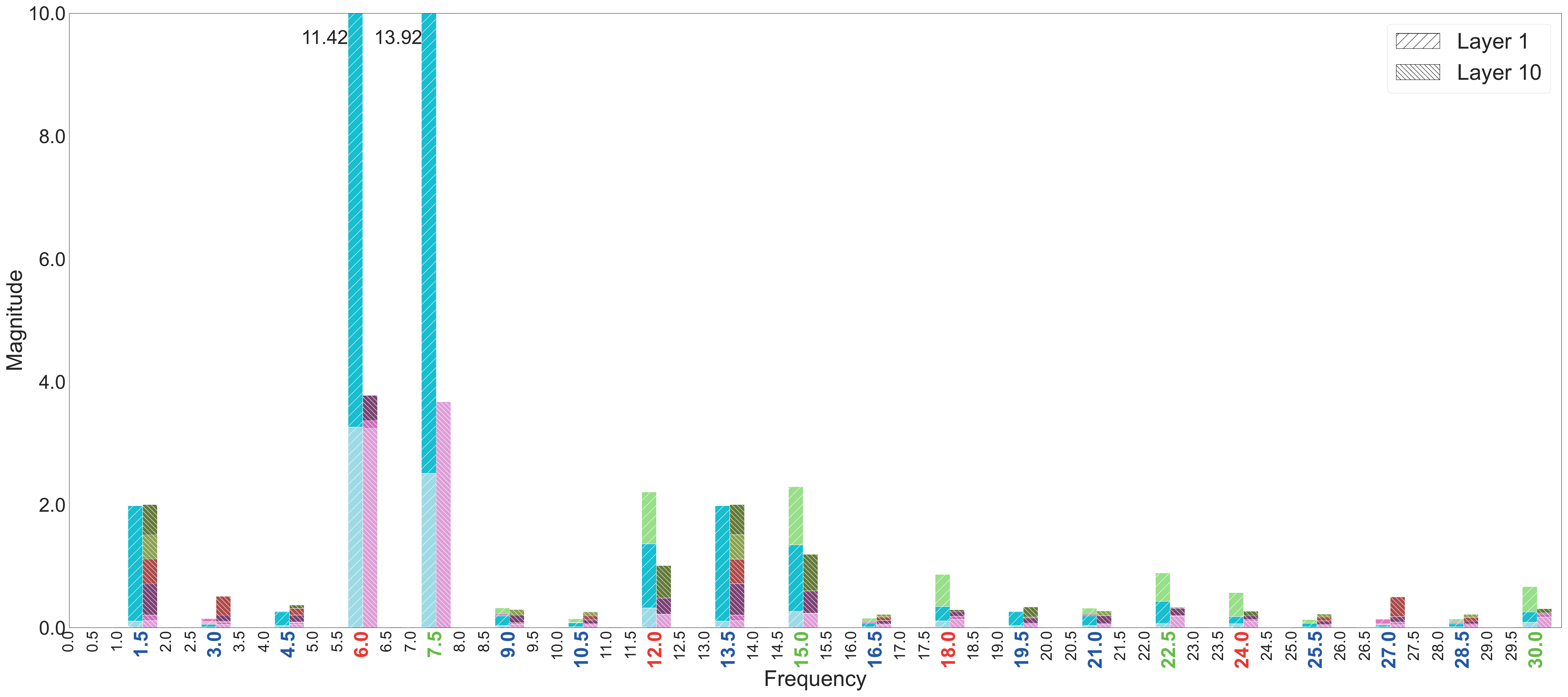}
  \end{center}
  \caption{The average frequencies across the 100 test images, obtained for the 16 filters of layers 1 and 10. Each color per bar corresponds to the magnitude of the frequency response for one of the filters of each of the layers. Along the x-axis: in red are the harmonics corresponding to the left-half of the images, in green the harmonics corresponding to the right-half of the images, and in blue are the intermodulation frequencies. Markers above some of the bars indicate frequency responses that surpass the chart's y-axis limit, which is set at 10 for visualization consistency.}
  \label{fftresponse}
\end{figure*}
Thus, 240 flickering images have been generated for each of the test images, and they are tagged at 6 Hz on their left half and at 7.5 Hz on their right half. Each 240-long image sequence has then been provided as input to the model, and then the mean value of each (convolutional) filter's response (i.e.~feature map) (excluding zeros) in the model is acquired and stored in the same order as that of the input

. And then, always in accordance with the procedure presented in Section \ref{sec:approach}, they are converted into the frequency domain.

Fig.~\ref{fftresponse} presents the average frequencies across the 100 test images, obtained for the 16 filters of layers 1 and 10. Each color per bar corresponds to the magnitude of the frequency response for one of the filters of each of the layers. The intervals between the bars, representing frequencies on the x-axis, are determined by 1/duration. In our case, with a duration of 2 seconds, this results in an interval of 0.5 Hz.

Notably, Fig.~\ref{fftresponse} reveals peaks at harmonics and intermodulation of the tagging frequencies, particularly up to the 4th-order frequency components. For $f_{left}=6$ Hz frequency, high peaks at the 1st, 2nd, 3rd, and 4th order harmonics - respectively as 6, 12, 18, and 24 Hz — are observed. Similarly, for the $f_{right}=7.5$ Hz frequency, we observe peaks at 7.5, 15, 22.5, and 30 Hz. Furthermore, we also identify peaks at intermodulation frequencies, including 2nd-order (1.5, 13.5), 3rd-order (4.5, 9, 19.5, 21), and 4th-order (3, 10.5, 16.5, 25.5, 27, 28.5). Importantly, for the other frequencies in the spectrum, the FFT values are very close to 0.

The peaks at harmonic frequencies affirm the neural network's capability to discern stimulus frequencies, showcasing its ability to generate distinct responses. Interestingly, the nodes that respond to the left half and right half can be discriminated by their magnitudes at harmonic frequencies. Furthermore, distinguishing nodes responding to the entire image is possible through their response to inter-modulation frequencies, especially if these nodes operate analogously to neurons in biological brains, particularly in the context of holistic representation \cite{boremanse2013objective}. 

\begin{figure*}[ht]
  \begin{center}
  \includegraphics[width=0.8\textwidth]{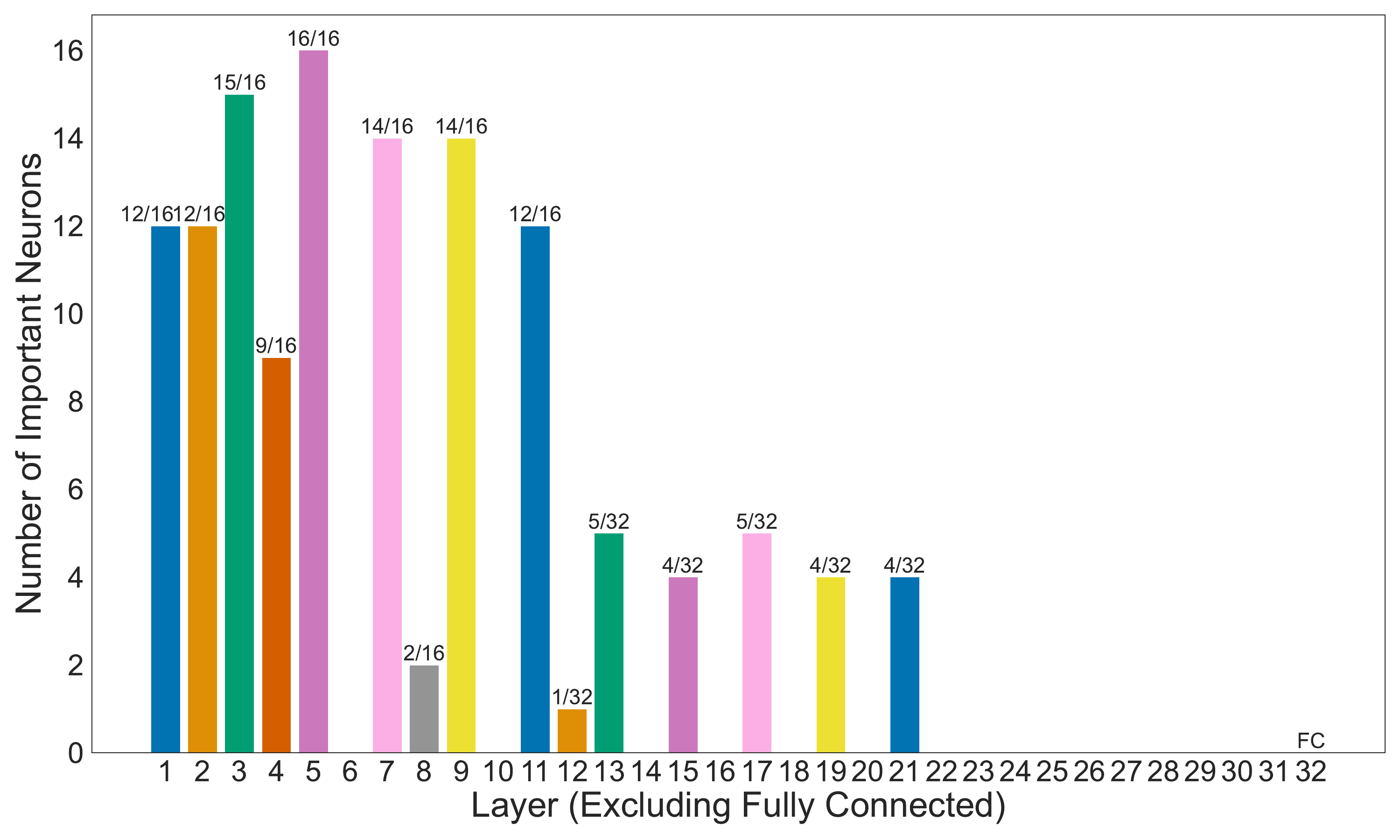}
  \end{center}
  \caption{Numbers of important filters per layer for the ResNet-32 according to the frequency tagging approach.}
  \label{numimportant}
\end{figure*}

Next, using an empirically determined SNR threshold of 150, the filters exhibiting SNR values above it across at least 50 of the 100 test images have been classified as important. Fig.~\ref{numimportant} shows the number of filters per layer that have been coined as important following this strategy. As we progress through the network layers towards the fully connected layer, there is a decrease in the number of filters exceeding the SNR threshold. This suggests that neurons in deeper layers contribute more evenly, making their individual impacts less distinguishable. The fully connected layer is excluded from this filter importance evaluation process due to its role in high-level decision-making, as it is less informative for the frequency-based analysis focused on visual feature extraction.

Although this experiment is by no means comprehensive, these preliminary results with a widely used CNN and a standard classification task, show that artificial neural networks (or at least the one tested) exhibit the same behavior of tuning to the flickering frequency of their input, thus paving the way for neuron/filter importance assessment through frequency tagging, similarly to the long-standing practice with biological brains. 

\section{Conclusion} 
\label{sec:conclusion}
In conclusion, this paper presents a novel approach to quantifying the significance of individual neurons within artificial neural networks (ANNs) by adopting the frequency tagging technique from neuroscience. By contrast-modulating (flickering) a network's input image in a sinusoidal fashion and observing the resulting neuron activations, the proposed method enables neuron-level analysis of a network's decision-making mechanisms. In our experiments with image classification, using a convolutional neural network (CNN), we observed notable harmonics as well as intermodulations when the resulting neuron-specific steady-state visually evoked potentials (SSVEP) are analyzed under part-based frequency tagging (left/right part: $f_1/f_2$). At each neuron, harmonics ($k f_1$) and intermodulations ($m f_1\pm nf_2$) quantify how much information from each input image part is conveyed and combined. We measure this through SNR which also draws the neuron's importance. Our findings suggest that ANNs exhibit similar behavior to biological brains in tuning to flickering frequencies, thus opening avenues for neuron/filter importance assessment through frequency tagging. This groundbreaking method offers potential applications in network pruning, and model interpretability, contributing to the advancement of explainable artificial intelligence (XAI) and addressing the lack of transparency in neural networks. One exciting future research is to develop novel loss functions that encourage the harmonics and intermodulations in the SSVEP spectrum. This would pull the network in a more biologically plausible manner and perhaps enable brain functions that have not been incorporated yet in ANNs.  

\bibliographystyle{unsrt}
\bibliography{ms}

\end{document}